\documentclass[a4paper]{IEEEtran}
\IEEEoverridecommandlockouts
\usepackage{amsmath,amssymb,bm,amsfonts,enumerate,url,footnote,color,ifpdf}
\usepackage[linesnumbered,ruled]
{algorithm2e}\usepackage[noend]{algpseudocode}
\ifCLASSINFOpdf
	\usepackage[pdftex]{graphicx}
         \usepackage{lipsum}
\else
	\usepackage[dvips]{graphicx}
         \usepackage{lipsum}
\fi
\usepackage{array,multirow,cite}
\ifCLASSOPTIONcompsoc
	\usepackage[caption=false,font=footnotesize,labelfont=sf,textfont=sf]{subfig}
\else
	\usepackage[caption=false,font=footnotesize]{subfig}
\fi
\usepackage[english]{babel}
\usepackage{booktabs,rotating,balance,arydshln} 
\usepackage{tikz} \usetikzlibrary{matrix}

\usepackage{titlesec}
\titlespacing*{\section}{0pt}{2.5ex plus 1ex minus .2ex}{1.3ex plus .2ex}

\usepackage{booktabs} 
\usepackage{tabularx} 

\begin{document}
\title{FedAuxHMTL: Federated Auxiliary Hard-Parameter Sharing Multi-Task Learning for Network Edge Traffic Classification}
\author{
    \IEEEauthorblockN{Faisal Ahmed\IEEEauthorrefmark{1}, Myungjin Lee\IEEEauthorrefmark{2}, Suresh Subramaniam\IEEEauthorrefmark{3}, Motoharu Matsuura\IEEEauthorrefmark{7}, Hiroshi Hasegawa\IEEEauthorrefmark{8}, and Shih-Chun Lin\IEEEauthorrefmark{1}} \\
    \IEEEauthorblockA{\IEEEauthorrefmark{1}North Carolina State University, Raleigh, NC, USA,
     \{fahmed5, slin23\}@ncsu.edu} \\
    \IEEEauthorblockA{\IEEEauthorrefmark{2}Cisco Research, Cisco Systems, Inc., CA, USA, 
    myungjle@cisco.com} \\
    \IEEEauthorblockA{\IEEEauthorrefmark{3}The George Washington University, Washington, USA, 
    suresh@gwu.edu} \\ 
    \IEEEauthorblockA{\IEEEauthorrefmark{7}University of Electro-Communications, Chofu, Japan, 
    m.matsuura@uec.ac.jp} \\
    \IEEEauthorblockA{\IEEEauthorrefmark{8}Nagoya University, Nagoya, Japan, 
    hasegawa@nuee.nagoya-u.ac.jp} \\
    
    \thanks{This work was supported in part by Cisco Research, the National Science Foundation (NSF) under Grant CNS-221034, Meta 2022 AI4AI Research, and the NC Space Grant.}
}

    

\maketitle
\thispagestyle{empty}

\begin{abstract}

Federated Learning (FL) has garnered significant interest recently due to its potential as an effective solution for tackling many challenges in diverse application scenarios, for example, data privacy in network edge traffic classification. Despite its recognized advantages, FL encounters obstacles linked to statistical data heterogeneity and labeled data scarcity during the training of single-task models for machine learning-based traffic classification, leading to hindered learning performance. In response to these challenges, adopting a hard-parameter sharing multi-task learning model with auxiliary tasks proves to be a suitable approach. Such a model has the capability to reduce communication and computation costs, navigate statistical complexities inherent in FL contexts, and overcome labeled data scarcity by leveraging knowledge derived from interconnected auxiliary tasks. This paper introduces a new framework for federated auxiliary hard-parameter sharing multi-task learning, namely, FedAuxHMTL. The introduced framework incorporates model parameter exchanges between edge server and base stations, enabling base stations from distributed areas to participate in the FedAuxHMTL process and enhance the learning performance of the main task—network edge traffic classification. Empirical experiments are conducted to validate and demonstrate the FedAuxHMTL's effectiveness in terms of accuracy, total global loss, communication costs, computing time, and energy consumption compared to its counterparts.
\end{abstract}

\IEEEpeerreviewmaketitle

\section{Introduction}
In recent years, federated learning (FL), an emerging decentralized machine learning (ML) technique, has gained significant attention within the research community. This heightened interest can be attributed to FL's ability to safeguard privacy and effectively tackle the obstacles linked to data sensitivity. In the context of FL, a diverse range of network components collaborate in the training of an ML model while preserving the decentralized nature of the training data \cite{9060868}. This strategy exhibits a notable contrast to other known strategies, such as the practice of uploading training data to a centralized server. However, The latter approach poses a considerable risk of data leakage and an increase in data communication overhead. Conversely, the model parameters derived from the ML model of local network components are combined and consolidated within the framework of FL. The aforementioned procedure results in the development of a globally robust model, hence enhancing privacy-preserving techniques through the provision of enhanced security and data access credentials \cite{9060868, 10138331}.

The FL approach endeavors to train a global model for a given task and is exclusively suitable for circumstances where the data is homogeneous\cite{10220110}. However, it is frequently seen that data distributions exhibit heterogeneity among various network components in numerous application scenarios. In this context, the network edge traffic classification may serve as an example where the allocation of uplink network traffic among multiple base stations (BSs) frequently illustrates non-independent and identically distributed (non-IID) characteristics. This heterogeneous non-IID attribute arises due to diverse constraints and requirements associated with these BSs.

Traffic classification is widely recognized as a fundamental task in the fields of network traffic monitoring, service improvement, and network security management. Specifically, it aims to establish a correlation between the network traffic and a certain category of traffic application, such as Google Voice, with the primary objective of guaranteeing quality of service (QoS) \cite{7557524}. Currently, ML methods are extensively utilized in traffic classification; for example, conventional ML techniques, such as KNN, Naïve Bayes, and SVM methods, are commonly employed in such classification. However, these methods require human efforts to design extracted features that are unable to apprehend the convoluted patterns of network traffic. Due to recent significant advancements in ML, deep learning is widely used in classification tasks \cite{9881547}. Specifically, it is favored for its exceptional ability to comprehend complex patterns and intelligently extract features. Nevertheless, deep learning requires a substantial quantity of labeled training data in a central location to learn patterns. Centralizing labeled training data incurs significant communication costs. In such a case, FL can be utilized to reduce the burden of high communication costs.

Moreover, the process of collecting and labeling traffic data is a time-intensive task \cite{9881547}. Furthermore, the limitations of obtaining labeled training data in network edge traffic classification combined with the FL data heterogeneity introduce additional obstacles. 
Multi-task learning (MTL) has emerged as an effective solution to combat such obstacles \cite{10220110,rezaei2020multitask}. Particularly, hard-parameter sharing MTL with auxiliary tasks can effectively addresses labeled data scarcity and data heterogeneity in FL scenarios. By integrating multiple auxiliary tasks, MTL enhances main task performance and improves computational efficiency compared to individual single-task models, as tasks can leverage shared information \cite{liebel2018auxiliary}.

Based on the aforementioned insights, we present \textbf{FedAuxHMTL}, i.e., \textbf{Fed}erated \textbf{Aux}iliary \textbf{H}ard-parameter sharing \textbf{M}ulti-\textbf{T}ask \textbf{L}earning, an auxiliary MTL-assisted FL approach tailored for network edge traffic classification that combats the aforementioned predicaments. The main contributions are summarized below:

\begin{itemize}


\item A two-tier auxiliary hard-parameter sharing MTL-supported edge server-base station (ES-BS) FL approach, namely, FedAuxHMTL, is introduced that aims to improve the main task (network edge traffic classification) performance in heterogeneous settings  characterized by limited labeled data. Additionally, this approach investigates the balance between computation and communication cost at the network edge, emphasizing privacy.

\item A 1D-CNN MTL model (Figure 1) is developed for simultaneously extracting detailed local features and capturing extensive long-distance correlation information in edge traffic, serving as the local classification model.

\item To improve the efficacy of traffic classification at the network edge more robustly, a random weighting strategy \cite{abs-2111-10603} is introduced to incorporate in FedAuxHMTL for the loss functions associated with auxiliary tasks.


\item The performance of FedAuxHMTL is assessed using a publicly available traffic dataset QUIC \cite{Rezaei2018HowTA} to showcase the superiority of the newly presented framework. Specifically, two auxiliary tasks, namely, bandwidth and network flow duration classification are derived from the QUIC dataset that contribute to enhancing the learning performance of the main task supported by FedAuxHMTL. Furthermore, practical simulations are carried out using the Cisco Flame platform \cite{101145}, facilitating insights for potential real-world deployment scenarios. Compared to the baselines, such as, FedAvg \cite{mcmahan2017communication}, the FedAuxHMTL exhibits greater gain, for example, 53\% and 32\% reduction in communication cost and computing time, respectively.

\begin{figure}
\centering
  \centerline{\includegraphics[width=\columnwidth]{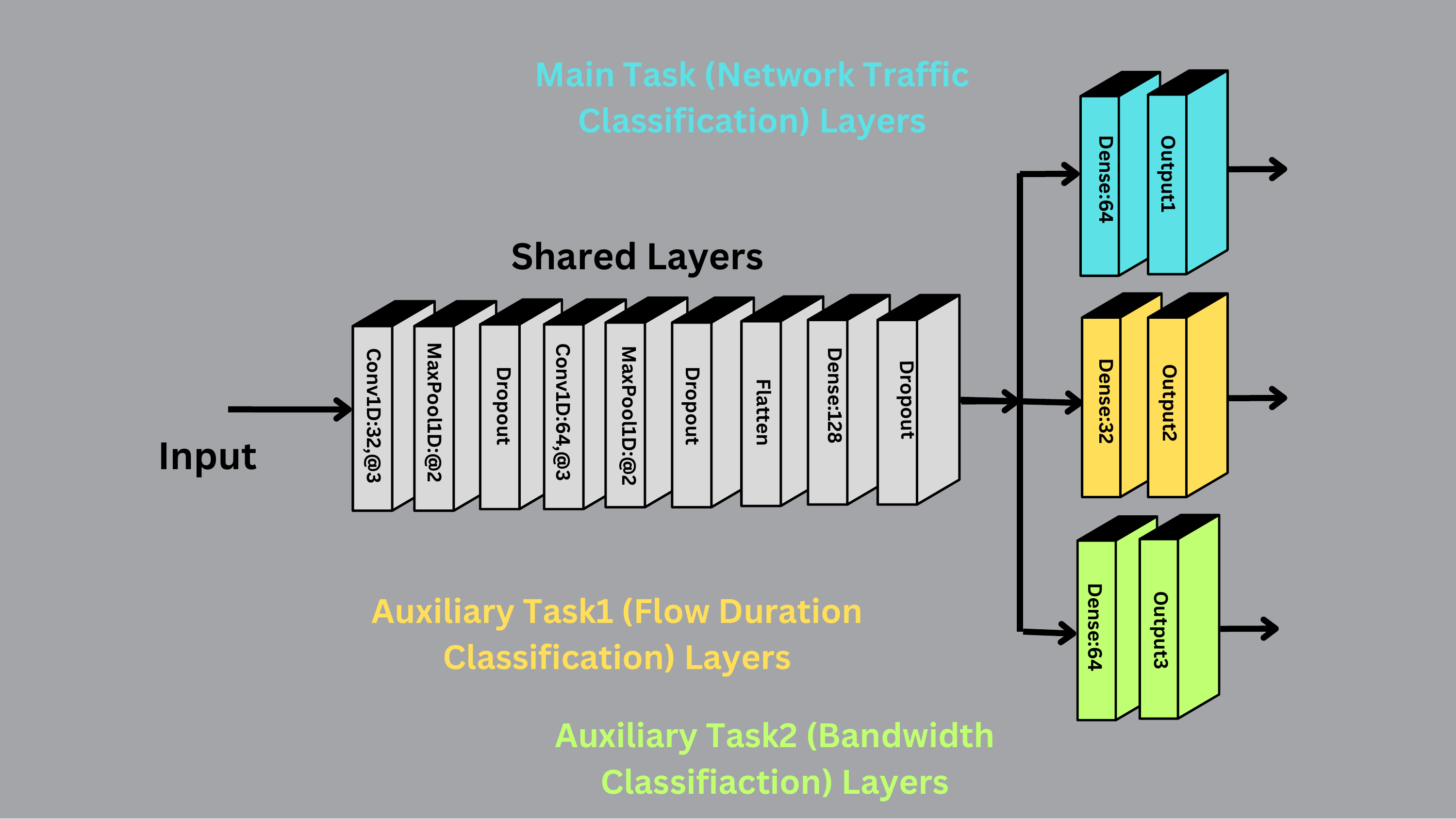}}
   \caption{1D-CNN MTL Local Model for FedAuxHMTL Framework.}
 \end{figure}
\end{itemize}

\label{sec:introduction}
\section{The State-of-the-Arts}

Deep learning, a prevalent ML technique has been widely applied to various classification challenges, including network traffic classification as highlighted in \cite{aceto2021distiller,electronics12173597,rezaei2020multitask}.
\cite{aceto2021distiller} introduces DISTILLER, an innovative multimodal MTL-deep learning approach designed to address the complex challenges of network traffic classification and enhance network visibility. \cite{electronics12173597} explores three MTL approaches to improve application traffic classification by exploiting task interdependencies, aiming to boost classification performance through the recognition of task relationships. The employment of MTL is advocated in \cite{rezaei2020multitask} for the simultaneous prediction of both bandwidth and duration of network traffic flows, as well as for classifying the network traffic into distinct categories. This study demonstrates that the prediction of bandwidth and flow duration can be achieved without the requirement for extensive labeling efforts or specific conditions, thereby rendering the need for ample training data.

Recently, in \cite{zhao2019multi}, authors introduced a multi-task deep neural network in FL, namely, MT-DNN-FL, for the concurrent execution of VPN traffic recognition, network anomaly detection, and network traffic classification tasks. According to their findings, the application of MT-DNN-FL results in a reduction of training time overhead in comparison to the several FL-DNN single-task models. The study in \cite{101007} presents a novel approach that combines MTL and FL to train personalized MTL models, namely, pFedDAMT, for encrypted traffic classification. The proposed pFedDAMT framework is designed to tackle the issue of data heterogeneity by employing a two-stage federated personalized MTL approach.

However, despite the effective approach used by the research conducted in \cite{aceto2021distiller}, \cite{electronics12173597}, and \cite{rezaei2020multitask} to address the challenges related to network edge traffic classification, the issue of privacy and security continues to persist due to the centralized settings. Although the investigations presented in \cite{zhao2019multi} and \cite{101007} employed FL settings as a remedy for these privacy and security issues, limitations still exist. Specifically, \cite{zhao2019multi} overlooked the aspect of data heterogeneity, while \cite{101007} acknowledged and addressed this concern, yet failed to take into account the  computation and communication costs. Moreover, both studies utilized conventional MTL to generalize all task performance rather than focusing solely on the performance improvement of the main task.

Contrary to the studies mentioned above, this study focuses on utilizing a hard-parameter sharing  MTL model, augmented with auxiliary tasks within FL settings. Specifically, this approach aimed at enhancing the learning performance of the main task while effectively managing data heterogeneity and mitigating the computation and communication costs.

\subsection{Learning Problem Explanations}

The primary goal of conventional FL methods, such as FedAvg \cite{mcmahan2017communication}, is to identify the stationary point $\Theta^*$ that minimizes the global loss function $\mathcal{L} (\mathsf{D}_{s},\Theta)$, mathematically: 
\begin{equation}
  \min_{\Theta \in \mathbb{R}^d} \mathcal{L}(\mathsf{D}_{s},\Theta),
\end{equation}
where, $\Theta$ $\in \mathbb{R}^d$ with $d$ denoted as the number of trainable model parameters. Moreover, $\mathcal{L} (\mathsf{D}_{s},\Theta)$ is the weighted average of $U_{s}$ BSs local loss function, can be denoted as:

\begin{equation}
\mathcal{L}(\mathsf{D}_{s},\Theta) \triangleq \frac{1}{\mathsf{D}_{s}} \sum_{u=1}^{U_s} \mathsf{D}_{u,s} \mathcal{L}_{u,s}(\mathsf{D}_{u,s},\Theta),
\end{equation}

In contrast to conventional FL, the MTL local loss function of the BS $\{u,s\}$ can be obtained by combining $M_{u,s}$ tasks $\mathcal{\{T_{\textit{i}}}\}_{i=1}^{M_{u,s}}$ as follows: 

\begin{equation}
  \mathcal{L}_{u,s}(\mathsf{D}_{u,s},\Theta) = \sum_{i=1}^{M_{u,s}} \mathcal{L}_{i}(\mathsf{D}_{u,s},\Theta) + \phi(\Theta,\mathsf{R}).
\end{equation} 
where $\phi(\cdot)$ is the regularization term that is designed to penalize the complexity of parameter weights. $\mathsf{R} \in \mathbb{R}^{m \times m}$ signifies the task relationship among all the tasks.  

However, in disparity to the above-mentioned conventional MTL that generalizes all tasks performance by optimizing the local loss function given in Eq. (3), the objective set for this study is to optimize the main task $l$ by leveraging the aid of an arbitrary number of related or auxiliary tasks $ b \in B_{u,s}$ in hard-parameter sharing MTL settings. Accordingly, Eq. (3) updated to \cite{101008}: 

\begin{equation}
  \mathcal{L}_{u,s}(\mathsf{D}_{u,s},\Theta) = \mathcal{L}_{l}(\mathsf{D}_{u,s},\Theta) + \sum_{b=1}^{B_{u,s}} \tau_{b} \mathcal{L}_{b}(\mathsf{D}_{u,s},\Theta).
\end{equation} 
where $\tau_{b}$ represents the weight assigned to auxiliary task $b$. It should be noted that the selection of $\tau_{b}$ can be either static, such as equal weighting, or dynamic, such as random weighting, depending on the decision made.

 \begin{figure} [!b]
  \centerline{\includegraphics[width=0.94\columnwidth]{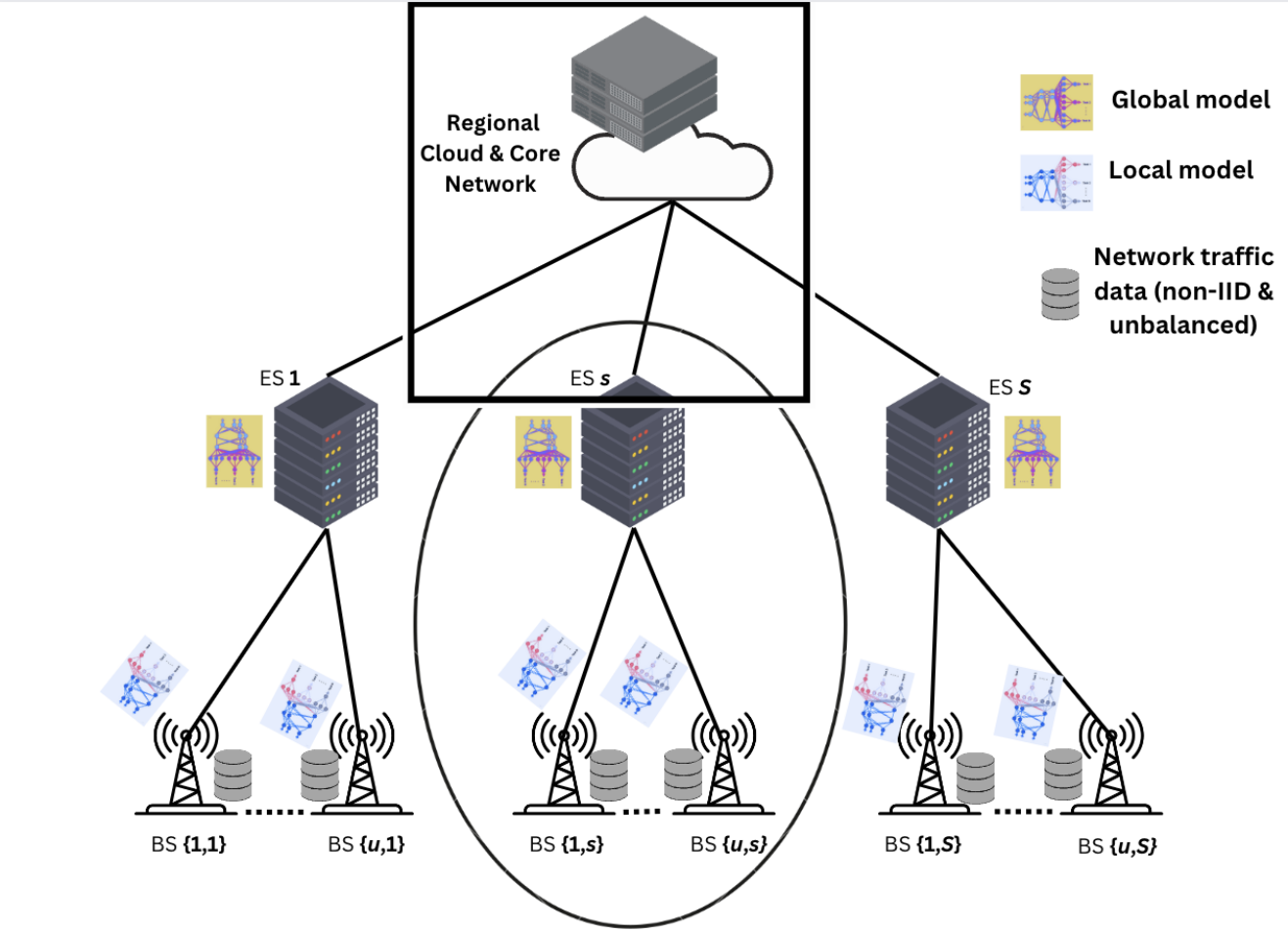}}
   \caption{Hard-parameter sharing federated multi-task learning in ES-BS.}
 \end{figure}

\section{System Model}
As shown in Figure 2, the introduced FedAuxHMTL consists of one region cloud, $S$ number of edge servers (ESs) that represented by set $\mathsf{S} = \{1,2,3,\dots,s,\dots,S\}$, and $U$ number of BSs that indexed by set $\mathsf{U} = \{1,2,3,\dots,U\}$. Let $\mathsf{U}_s = \{1,2,3,\dots,u,\dots,U_{s}\}$ denote the set of BSs covered by the ES $s$. BS $\{u,s\}$ denotes that the $u$-th BS is served by the ES $s$, where $u\in \mathsf{U_s}$.  It is assumed that BS $\{u,s\}$ contains a dataset $\mathsf{D}_{u,s}$ where $D_{u,s}$ is its total number of data samples. It is also assumed that there are no overlapping among datasets from different BSs that are covered by the ES $s$, thus, $\mathsf{D}_{p,s} \cap \mathsf{D}_{u,s} = \emptyset, (\forall p,u \in \mathsf{U}_s, s \in \mathsf{S})$. Additionally, the whole dataset for the ES $s$ can be denoted as $\mathsf{D}_s = \cup \{\mathsf{D}_{u,s}\}^{U_s}_{u=1}$. In a similar manner, the total number of training data samples for the ES $s$ can be defined as ${D}_s =\sum_{u=1}^{U_s} {D}_{u,s}$. The $\mathsf{D}_s$ exhibits non-IID characteristics across $U_{s}$ BSs over the entire duration of communication rounds $T$.

\subsection{Local Computation \& Energy Consumption Model}
This study examines a local computation model that is comparable to the one described in \cite{9629331}. Let $C_{u,s}$ represent the amount of CPU cycles required by BS $\{u,s\}$ to handle per bit of data sample. It is assumed that the size of all data samples are fixed. Therefore, to compute one local iteration, the required CPU cycles can be calculated as $C_{u,s} {{L}_{u,s}} \cdot f_{u,s}$, where, $f_{u,s}$ denotes the frequency of required CPU cycles and ${L}_{u,s}$ denotes the length of local dataset $\mathsf{D}_{u,s}$. In addition, the time required for computing one local iteration of BS $\{u,s\}$ can be calculated as follows:

\begin{equation}
t^{loc}_{u,s} = \frac{C_{u,s} {{L}_{u,s}}}{f_{u,s}}.
\end{equation} 
 
It should be noted that the computational speed of a CPU is often quantified by its frequency. In practice, the adjustment of the frequency of CPU cycles is undertaken with the aim of mitigating energy consumption and preventing device overheating. Therefore, the energy consumption of CPU for the computation of one local iteration in BS $\{u,s\}$ can be expressed as \cite{Burd1996ProcessorDF}: 

\begin{equation}
 E^{comp}_{u,s} = 0.5{\beta}C_{u,s}{L}_{u,s}f_{u,s}^2,
\end{equation}
 where, $0.5{\beta}$ is the effective coefficient capacitance of BS $\{u,s\}$'s computing chip.

 In a similar manner, if there are required $I_{s}$ local iterations to reach a targeted accuracy level $\kappa$, then, the energy consumption of each BS $\{u,s\}$ can be derived as:

\begin{equation}
 E^{comp}_{{u,s}_{total}} = I_{s}E^{comp}_{u,s} = 0.5{\beta}I_{s}C_{u,s}{L}_{u,s}f_{u,s}^2.
\end{equation}

\subsection{Communication Costs Model} 

 There are two communication phases associated with FedAuxHMTL, namely, the uplink communication phase and the downlink communication phase. Lets assume ${P}_s^{t} ({P}_s^{t} \subseteq {U}_s)$ number of BSs  engage in the exchange of model parameters during the $t$-th communication round. By making the assumption of a homogeneous environment, where each BS $\{u,s\}$ shares the same model parameters denoted as $\Theta$, with the number of model parameters being identical for all BSs, and the total communication round $T$, the total communication costs $\omega_s$ associated with the ES $s$ and $U_{s}$ number of BSs can be determined as follows \cite{9505307}: 

 \begin{equation}
 \omega_s = \sum_{t=1}^T \{{P}_s^{t} + {U}_s\} \cdot \Theta,
 \end{equation}

Similarly, if a targeted test accuracy level $\kappa$ requires $\varkappa$ communication rounds, then, the communication costs, $\omega_{s}^{\kappa}$ can be calculated as follows:  

\begin{equation}
\omega_{s}^{\kappa} = \sum_{\iota=1}^{\varkappa} \{{P}_s^{\iota} + {U}_s\} \cdot \Theta.
\end{equation}

 \section{FedAuxHMTL: Federated Auxiliary Hard-parameter sharing Multi-Task Learning}
 
 The introduced FedAuxHMTL framework operates in an iterative manner by performing the following steps during each communication round $t$.

\begin{itemize}
\item \textit{Step} 1: The ES $s$ broadcasts the current global model parameters $\Theta^{t}$ to the $U_{s}$ BSs that have been involved in the learning process and are within its coverage area. 

\item \textit{Step} 2: Each BS $\{u,s\}$ collects a batch sample of length $\upsilon$ from its local dataset $\mathsf{D}_{u,s}$ and applies the stochastic gradient descent (SGD) algorithm or a variant thereof to locally update the received global model parameters $\Theta^{t}$, mathematically:

\begin{equation}
\Theta^{t}_{u,s} \leftarrow \Theta^{t}_{u,s} - \eta \nabla\mathcal{L}_{u,s}(\mathsf{D}_{u,s},\Theta^{t}_{u,s}).
\end{equation}
where $\eta$ is the gradient descent step size.

\item \textit{Step} 3: After several rounds of local updates, each BS $\{u,s\}$ transmits the updated $\Theta^{t}_{u,s}$ to the ES $s$.

\item \textit{Step} 4: Upon receiving the updated local model parameters from each BS $\{u,s\}$, the ES $s$ employs a global aggregation technique, such as the one proposed in \cite{mcmahan2017communication}, to combine all the updated local model parameters and obtain the subsequent global model parameter $\Theta^{t+1}$, mathematically:

\begin{equation}
  \Theta^{t+1} = \frac{1}{\mathsf{D}_{s}} \sum_{u=1}^{U_s} \mathsf{D}_{u,s} \Theta^{t}_{u,s}.
\end{equation}
The model parameters, which have been aggregated, are then transmitted back to each BS $\{u,s\}$ in order to be used in the next training communication round. The aforementioned process is iterated until it reaches the predetermined maximum number of iterations.
\end{itemize}

\subsection{Random Loss Weighting}

 In the realm of auxiliary hard-parameter sharing MTL, where the main task's performance is prioritized, a significant challenge is balancing the auxiliary tasks. Various dynamic loss weighting strategies have been developed to mitigate this issue. A notable example, as discussed in \cite{abs-2111-10603}, is the random loss weighting strategy. This approach involves assigning loss weights randomly from a distribution to train a hard-parameter sharing MTL model. This study adopts the aforementioned strategy to optimize loss weights for auxiliary tasks aiming to improve the main task's learning performance.        

In the current study the non-negativity of loss weights in $\tau$ for auxiliary tasks is ensured through a two-step process. Firstly, a vector $\widetilde{\tau}$ = $(\widetilde{\tau_{1}}, \dots, \widetilde{\tau}_{{B_{u,s}}})$ is obtained by sampling from a distribution $p(\widetilde{\tau})$. Then, $\widetilde{\tau}$ is normalized into $\tau$ using a mapping $\gamma$, where $\gamma : \mathbb{R}^{B_{u,s}} \rightarrow \Delta^{B_{u,s}-1}$ is a normalization function. The normalization function ensures that $\tau$ satisfies the conditions of being a simplex in $\mathbb{R}^{B_{u,s}}$, i.e., $\tau \in \mathbb{R}^{B_{u,s}}$ with $\sum^{B_{u,s}}_b \tau_{b} = 1$ and $\tau_{b} \geq 0$. Notably, the Softmax function is adopted as the normalization function in the study, mathematically:

\begin{equation}
\mathbb{E}[\tau] = \gamma(\widetilde{\tau})_n = \frac{e^{\widetilde{\tau}_n}}{\sum^{N}_{j=1} e^{\widetilde{\tau}_j}}.
\end{equation}
where, $\mathbb{E}[\tau]$ denotes as the expectation of loss weights $\tau$ (random variables) over the entire training iterations.

\begin{figure*} [!t]
\centerline{\includegraphics[width=2.12\columnwidth]{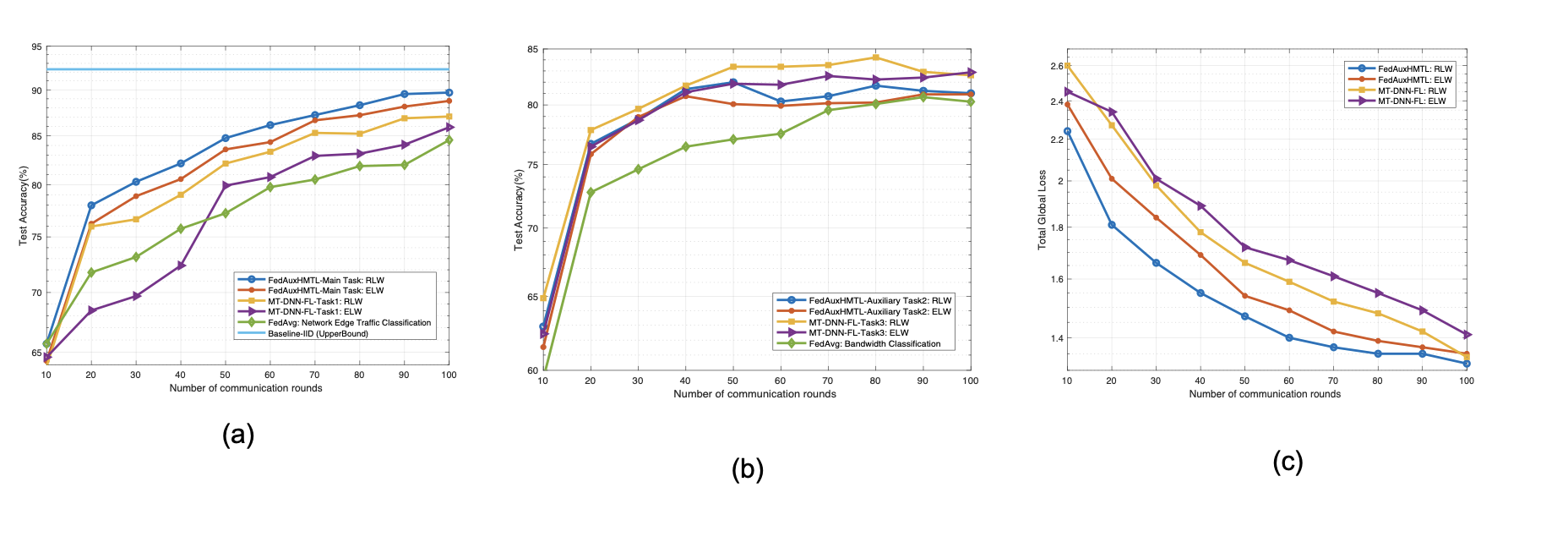}}
\caption{(a) Test Accuracy vs the Number of Communication Rounds for the Main Task and Baselines, (b) Test Accuracy vs the Number of Communication Rounds for the Auxiliary Task2 and Baselines, (c) Total Global Loss vs the Number of Communication Rounds.}
\end{figure*}

\section{Performance Evaluation}

\subsection{Simulation Setup}
\subsubsection{Dataset}
The QUIC dataset \cite{Rezaei2018HowTA} is introduced in this study as the dataset for network edge traffic classification. The dataset comprises five distinct Google services, namely, Google Drive (1664 flows), Google Search (1945 flows), Google Music (622 flows), Google Doc (1251 flows), and Youtube (1107 flows). 

The initial dataset is consisting of 6439 data samples and partitioned into two subsets: 90$\%$ of the data samples were allocated for training purposes, while the remaining 10$\%$ were reserved for testing. Subsequently, 90$\%$ of the training data samples are allocated to all BSs in a non-IID manner. In addition to this, 10$\%$ of the allocated training samples were utilized for validation.


\subsubsection{Simulation Platform}

The FedAuxHMTL is implemented on the CISCO Flame platform \cite{101145}. The Flame is a comprehensive tool for customizing FL applications to fit specific deployment contexts. Moreover, Flame's key feature is its system-level configuration capability for FL application topologies, which is further enhanced by its scalability to new FL architectures. This customization is facilitated through Topology Abstraction Graphs (TAGs), which separate ML application logic from deployment complexities. This approach significantly reduces the need for development resources in tailoring application deployment.

In the simulation, 6 BSs are considered that are covered by ES $s$ and each BS $\{u,s\}$ optimizes its performance by employing the SGD optimizer with a learning rate, $\eta$ = 0.005. A consistent BS batch size, $\sigma$ = 32 and a local epoch, $\varrho$ = 20 are employed in $T$ communication rounds. During each communication round $t$, all BSs participate in the training process. Besides, $\beta$ = 2$\times 10^{-28}$, $C_{u,s}$ = 40 cycles/bit, and $f_{u,s}$ = 2.0 GHz are assumed. The total count of trainable parameters is 257,358. The Tensorflow-Keras platform is utilized for the construction of 1D-CNN. The simulations were conducted on a 2021 MacBook Pro equipped with an Apple M1 Pro chip, featuring an 8-core CPU, a 14-core GPU, and 16GB of unified RAM.

\subsubsection{Performance Metrics}
\begin{itemize}

\item \textbf{Accuracy}: Accuracy is defined as the ratio of the number of correct predictions to the total number of predictions.
\item \textbf{Computing Time}: The amount of time required to reach a particular number of communication rounds, $\varkappa$.
\item \textbf{Energy Consumption}: The amount of energy expenditure to reach a specific level of test accuracy, $\kappa$.

\end{itemize}

\subsection{Baseline Benchmarks}
\begin{itemize}

\item \textbf{Baseline-IID}: This approach involves utilizing 1D-CNN for auxiliary MTL in an IID environment with FL setting and considered as the UpperBound for this study. 
\item \textbf{FedAvg \cite{mcmahan2017communication}}: The FL framework trains a single global model for a specific task.
\item \textbf{MT-DNN-FL \cite{zhao2019multi}}: A conventional MTL model that is employed in FL framework to generalize all three tasks performance. In this study, for MT-DNN-FL, three tasks are assigned as follows: \textbf{Task 1:} \textit{network edge traffic classification}, \textbf{Task 2:} \textit{flow duration classification}, \textbf{Task 3:} \textit{bandwidth classification}. 

\end{itemize}


\begin{table}[b]
\centering
\caption{Comparative Analysis of Energy Consumption and Communication Cost between \textbf{FedAuxHMTL-Main Task} and Baselines: $\kappa = 80\%, \varrho = 20$ (\textbf{RLW = Random Loss Weighting, ELW = Equal Loss Weighting})}
\small 
\begin{tabularx}{\columnwidth}{Xcc}
\toprule
& \multicolumn{1}{c}{\textbf{Energy}} & \multicolumn{1}{c}{\textbf{Communication}} \\
& \multicolumn{1}{c}{\textbf{Consumption (J)}} & \multicolumn{1}{c}{\textbf{Cost (MB)}} \\
\midrule
FedAuxHMTL: RLW & \textbf{76.44} & \textbf{349.91} \\
FedAuxHMTL: ELW & 97.53 & 446.44 \\
MT-DNN-FL: RLW & 115.98 & 530.90 \\
MT-DNN-FL: ELW & 150.24 & 687.76 \\
FedAvg & 163.42 & 748.02 \\
\bottomrule
\end{tabularx}
\end{table}

\begin{table}[ht]
\centering
\caption{Comparative Analysis of Energy Consumption and Communication Cost between (\textbf{FedAuxHMTL-Auxiliary Task 1} and Baselines: $\kappa = 75\%, \varrho = 20$) \& (\textbf{FedAuxHMTL-Auxiliary Task 2} and Baselines: $\kappa = 80\%, \varrho = 20$)}
\small 
\begin{tabularx}{\columnwidth}{Xcc}
\toprule
& \multicolumn{1}{c}{\textbf{Energy}} & \multicolumn{1}{c}{\textbf{Communication}} \\
& \multicolumn{1}{c}{\textbf{Consumption (J)}} & \multicolumn{1}{c}{\textbf{Cost (MB)}} \\
\midrule
FedAuxHMTL: RLW & 55.35 & 253.38 \\
FedAuxHMTL: ELW & 68.54 & 313.71 \\
MT-DNN-FL: RLW & \textbf{52.71} & \textbf{241.32} \\
MT-DNN-FL: ELW & 65.90 & 301.65 \\
FedAvg & 65.90 & 301.65 \\
\bottomrule
\bottomrule
FedAuxHMTL: RLW & 92.25 & 422.31 \\
FedAuxHMTL: ELW & 89.62 & 410.24 \\
MT-DNN-FL: RLW & \textbf{81.74} & \textbf{374.04} \\
MT-DNN-FL: ELW & 89.62 & 410.24 \\
FedAvg & 210.87 & 965.27 \\
\bottomrule
\end{tabularx}
\end{table}

\begin{table}[ht]
\centering
\caption{Computing Time of \textbf{FedAuxHMTL} and Baselines: $\varkappa$ = 100}
\small 
\begin{tabularx}{\columnwidth}{Xcc}
\toprule
& \textbf{Computing Time (sec)} \\
\midrule
FedAuxHMTL: RLW & 2242.36 \\
FedAuxHMTL: ELW & \textbf{2054.66} \\
MT-DNN-FL: RLW & 2263.59 \\
MT-DNN-FL: ELW & \textbf{2049.15} \\
FedAvg-Combined & 3032.54 \\
\bottomrule
\end{tabularx}
\end{table}

\subsection{Simulation Results}
From Figure 3(a) it is depicted that FedAuxHMTL with RLW attains the highest test accuracy compared to the baselines for $\varkappa$ = 100. More specifically, to reach a targeted accuracy, such as, $\kappa$ = 80$\%$ for the network edge traffic classification (main task), FedAuxHMTL with RLW and FedAuxHMTL with ELW took 29 and 37 communication rounds, respectively, whereas, baselines, such as, FedAvg, conventional FMTL with RLW, and conventional FMTL with ELW took 62, 44, and 57 communication rounds to reach the targeted test accuracy, respectively. Hence, it can be conclude that FedAuxHMTL has higher convergence speed compared to the baselines. The reason behind this is that, due to using auxiliary MTL setup in FedAuxHMTL, communication efficacy improved as well as privacy is preserved. It should be noted that in an IID environment with FL setting the accuracy reaches about 92$\%$ for $\varkappa$ = 100 which is the UpperBound evaluated in this work. 


Figures 3(b) and 3(c) shows test accuracies for FedAuxHMTL and baseline tasks, with targeted test accuracy $\kappa$ set to 75\% and 80\% for baseline task 2 and 3, respectively. Notably, baseline MT-DNN-FL with RLW and ELW outperformed others, requiring the lowest and second-lowest number of communication rounds for $\kappa$, respectively. In contrast, FedAuxHMTL when employing RLW and ELW for auxiliary tasks required lower number of communication rounds than FedAvg only; the reason is, the loss function of auxiliary tasks is modulated by weight factors, effectively elevating the importance of the main task relative to the auxiliary tasks.

Figure 3(d) illustrates the total global loss of FedAuxHMTL. This particular result is included to demonstrate that FedAuxHMTL can achieve convergence for non-convex loss functions. Furthermore, it is evident that FedAuxHMTL, when utilizing RLW and ELW exhibits the lowest overall global loss in comparison to the baselines. The reason is, due to use of MTL framework with auxiliary tasks, the main task got the lowest loss and consequently, this led to a reduction in the total global loss as compared to conventional FMTL. 

Examining Figure 3(a), it is evident that FedAuxHMTL with RLW required the lowest number of communication rounds to attain the targeted test accuracy, $\kappa$ = 80$\%$, while FedAvg needed the highest number of communication rounds. Notably, to achieve $\kappa$, FedAuxHMTL with RLW exhibits the lowest communication cost and energy consumption, whereas FedAuxHMTL with ELW incurs slightly higher communication cost and energy consumption for the main task. In contrast, the baselines exhibit higher communication cost and energy consumption as depicted in Table I. The observed behavior can be attributed to factors described in Figure 3(a).

Table II demonstrates that the baseline MT-DNN-FL for Task 2 and Task 3 exhibits the lowest energy consumption and communication cost relative to other frameworks. The underlying reasons for this are elucidated in Figures 3(b) and 3(c).
 
Table III shows the computing times for FedAuxHMTL compared to baseline models. For $\varkappa$ = 100, FedAvg's computing time for Task 1, Task 2, and Task 3 are 1043.45 sec, 1079.48 sec, and 909.61 sec, totaling 3032.54 sec for FedAvg-Combined, significantly higher than FedAuxHMTL with ELW. This difference stems from the simpler network structure of FedAuxHMTL and MT-DNN-FL, since FedAvg-Combined cannot share layers between tasks. FedAuxHMTL with RLW has slightly longer computing time than with ELW, due to the Softmax function's exponential component in its weighting strategy. A similar explanation applies to MT-DNN-FL, due to its structure for sharing parameters.

\section{Conclusion and Future works}
In this paper, we introduced FedAuxHMTL that significantly improves the learning performance of network edge traffic classification. Additionally, FedAuxHMTL demonstrates its effectiveness in mitigating communication and computation costs. The equal weighting and the designed random weighting are integrated with the introduced FedAuxHMTL and compacted with existing federated multi-task and single-task learning approaches. Simulation results showed and validated the efficacy of the introduced FedAuxHMTL framework. 

For future work, we will provide a comprehensive theoretical analysis among the latest FMTL solutions. 




\bibliographystyle{IEEEtran}
\bibliography{IEEEabrv,bib}

\begin{thebibliography}{10}
\providecommand{\url}[1]{#1}
\csname url@samestyle\endcsname
\providecommand{\newblock}{\relax}
\providecommand{\bibinfo}[2]{#2}
\providecommand{\BIBentrySTDinterwordspacing}{\spaceskip=0pt\relax}
\providecommand{\BIBentryALTinterwordstretchfactor}{4}
\providecommand{\BIBentryALTinterwordspacing}{\spaceskip=\fontdimen2\font plus
\BIBentryALTinterwordstretchfactor\fontdimen3\font minus \fontdimen4\font\relax}
\providecommand{\BIBforeignlanguage}[2]{{%
\expandafter\ifx\csname l@#1\endcsname\relax
\typeout{** WARNING: IEEEtran.bst: No hyphenation pattern has been}%
\typeout{** loaded for the language `#1'. Using the pattern for}%
\typeout{** the default language instead.}%
\else
\language=\csname l@#1\endcsname
\fi
#2}}
\providecommand{\BIBdecl}{\relax}
\BIBdecl

\bibitem{9060868}
W.~Y.~B. Lim, N.~C. Luong, D.~T. Hoang, Y.~Jiao, Y.-C. Liang, Q.~Yang, D.~Niyato, and C.~Miao, ``Federated learning in mobile edge networks: A comprehensive survey,'' \emph{IEEE Communications Surveys and Tutorials}, vol.~22, no.~3, pp. 2031--2063, 2020.

\bibitem{10138331}
S.-C. Lin, C.-H. Lin, and M.~Lee, ``Privacy-preserving serverless edge learning with decentralized small data,'' \emph{IEEE Network}, pp. 1--8, 2023.

\bibitem{10220110}
H.~Zhang, M.~Tao, Y.~Shi, X.~Bi, and K.~B. Letaief, ``Federated multi-task learning with non-stationary and heterogeneous data in wireless networks,'' \emph{IEEE Transactions on Wireless Communications}, pp. 1--1, 2023.

\bibitem{7557524}
P.~Wang, S.-C. Lin, and M.~Luo, ``A framework for qos-aware traffic classification using semi-supervised machine learning in sdns,'' in \emph{2016 IEEE International Conference on Services Computing (SCC)}, 2016, pp. 760--765.

\bibitem{9881547}
Y.~Guo and D.~Wang, ``Feat: A federated approach for privacy-preserving network traffic classification in heterogeneous environments,'' \emph{IEEE Internet of Things Journal}, vol.~10, no.~2, pp. 1274--1285, 2023.

\bibitem{rezaei2020multitask}
S.~Rezaei and X.~Liu, ``Multitask learning for network traffic classification,'' in \emph{29th International Conference on Computer Communications and Networks (ICCCN)}, 2020.

\bibitem{liebel2018auxiliary}
L.~Liebel and M.~K{\"o}rner, ``Auxiliary tasks in multi-task learning,'' \emph{arXiv preprint arXiv: 1805.06334}, 2018.

\bibitem{abs-2111-10603}
B.~Lin, F.~Ye, and Y.~Zhang, ``A closer look at loss weighting in multi-task learning,'' \emph{CoRR}, vol. abs/2111.10603, 2021.

\bibitem{Rezaei2018HowTA}
S.~Rezaei and X.~Liu, ``How to achieve high classification accuracy with just a few labels: A semi-supervised approach using sampled packets,'' \emph{ArXiv}, vol. abs/1812.09761, 2018.

\bibitem{101145}
H.~Daga, J.~Shin, D.~Garg, A.~Gavrilovska, M.~Lee, and R.~R. Kompella, ``Flame: Simplifying topology extension in federated learning,'' in \emph{Proceedings of the 2023 ACM Symposium on Cloud Computing}, ser. SoCC '23.\hskip 1em plus 0.5em minus 0.4em\relax New York, NY, USA: Association for Computing Machinery, 2023, p. 341–357.

\bibitem{mcmahan2017communication}
B.~McMahan, E.~Moore, D.~Ramage, S.~Hampson, and B.~A. y~Arcas, ``Communication-efficient learning of deep networks from decentralized data,'' in \emph{Artificial intelligence and statistics}.\hskip 1em plus 0.5em minus 0.4em\relax PMLR, 2017, pp. 1273--1282.

\bibitem{aceto2021distiller}
G.~Aceto, D.~Ciuonzo, A.~Montieri, and A.~Pescap{\'e}, ``Distiller: Encrypted traffic classification via multimodal multitask deep learning,'' \emph{Journal of Network and Computer Applications}, vol. 183, p. 102985, 2021.

\bibitem{electronics12173597}
U.-J. Baek, B.~Kim, J.-T. Park, J.-W. Choi, and M.-S. Kim, ``A multi-task classification method for application traffic classification using task relationships,'' \emph{Electronics}, vol.~12, no.~17, 2023.

\bibitem{zhao2019multi}
Y.~Zhao, J.~Chen, D.~Wu, J.~Teng, and S.~Yu, ``Multi-task network anomaly detection using federated learning,'' in \emph{Proceedings of the 10th international symposium on information and communication technology}, 2019, pp. 273--279.

\bibitem{101007}
X.~Guan, R.~Du, X.~Wang, and H.~Qu, ``A personalized federated multi-task learning scheme for encrypted traffic classification,'' in \emph{Artificial Neural Networks and Machine Learning -- ICANN 2023}.\hskip 1em plus 0.5em minus 0.4em\relax Cham: Springer Nature Switzerland, 2023, pp. 258--270.

\bibitem{101008}
Z.~Zhang, P.~Luo, C.~C. Loy, and X.~Tang, ``Facial landmark detection by deep multi-task learning,'' in \emph{Computer Vision -- ECCV 2014}, D.~Fleet, T.~Pajdla, B.~Schiele, and T.~Tuytelaars, Eds.\hskip 1em plus 0.5em minus 0.4em\relax Cham: Springer International Publishing, 2014, pp. 94--108.

\bibitem{9629331}
J.~Feng, L.~Liu, Q.~Pei, and K.~Li, ``Min-max cost optimization for efficient hierarchical federated learning in wireless edge networks,'' \emph{IEEE Transactions on Parallel and Distributed Systems}, vol.~33, no.~11, pp. 2687--2700, 2022.

\bibitem{Burd1996ProcessorDF}
T.~D. Burd and R.~W. Brodersen, ``Processor design for portable systems,'' \emph{Journal of VLSI signal processing systems for signal, image and video technology}, vol.~13, pp. 203--221, 1996.

\bibitem{9505307}
S.~Liu, J.~Yu, X.~Deng, and S.~Wan, ``Fedcpf: An efficient-communication federated learning approach for vehicular edge computing in 6g communication networks,'' \emph{IEEE Transactions on Intelligent Transportation Systems}, vol.~23, no.~2, pp. 1616--1629, 2022.

\end{thebibliography}
\end{document}